\documentclass[letterpaper, 10 pt, conference]{ieeeconf}  

\IEEEoverridecommandlockouts                              

\usepackage{graphics} 
\usepackage{epsfig} 
\usepackage{mathptmx} 
\usepackage{times} 
\usepackage{amsmath} 
\usepackage{amssymb}  
\usepackage{booktabs}
\usepackage{multirow}
\usepackage{xspace}
\usepackage{hyperref}
\usepackage{cleveref}
\usepackage{caption}
\usepackage{subcaption}
\usepackage{float}
\usepackage{svg}
\usepackage{cite}
\usepackage[T1]{fontenc}

\crefname{algorithm}{Alg.}{Algs.}
\Crefname{algocf}{Algorithm}{Algorithms}
\crefname{section}{Sec.}{Secs.}
\Crefname{section}{Section}{Sections}
\crefname{table}{Tab.}{Tabs.}
\Crefname{table}{Table}{Tables}
\crefname{figure}{Fig.}{Fig.}
\Crefname{figure}{Figure}{Figure}
\crefname{appendix}{Appx.}{Appx.}
\Crefname{appendix}{Appendix}{Appendix}

\usepackage{tikz}
\usepackage{tabularx}
\usepackage{adjustbox}

\newcommand{\tick}{\textcolor{green!70!black}{\checkmark}}
\newcommand{\cross}{\textcolor{red}{X}}

\newcommand{\env}{\emph{OmniDrones}\xspace}

\newcommand{\drone}[1]{\textit{#1}}

\title{\LARGE \bf
\textit{OmniDrones}: An Efficient and Flexible Platform for Reinforcement Learning in Drone Control
}

\author{Botian Xu$^{1*}$,
Feng Gao$^{1}$,
Chao Yu$^{1\sharp}$,
Ruize Zhang$^{1}$,
Yi Wu$^{12}$,
Yu Wang$^{1\sharp}$
\thanks{$^{\sharp}$ Corresponding Author}
\thanks{$^1$ Tsinghua University $^2$ Shanghai Qi Zhi Institute}
\thanks{$^{*}$ Work done as an intern in Tsinghua University}
\thanks{This work has been submitted to the IEEE for possible publication. Copyright may be transferred without notice, after which this version may no longer be accessible}
}

\begin{document}

\maketitle
\thispagestyle{empty}
\pagestyle{empty}


\begin{abstract}


In this work, we introduce \env, an efficient and flexible platform tailored for reinforcement learning in drone control, built on Nvidia's Omniverse Isaac Sim. It employs a bottom-up design approach that allows users to easily design and experiment with various application scenarios on top of GPU-parallelized simulations. It also offers a range of benchmark tasks, presenting challenges ranging from single-drone hovering to over-actuated system tracking. In summary, we propose an open-sourced drone simulation platform, equipped with an extensive suite of tools for drone learning. It includes 4 drone models, 5 sensor modalities, 4 control modes, over 10 benchmark tasks, and a selection of widely used RL baselines. To showcase the capabilities of \env and to support future research, we also provide preliminary results on these benchmark tasks. We hope this platform will encourage further studies on applying RL to practical drone systems. For more resources including documentation and code, please visit: \url{https://omnidrones.readthedocs.io/}.

\end{abstract}

\section{INTRODUCTION}

Multi-rotor drones and multi-drone systems are receiving increasing attention from both industry and academia due to their remarkable agility and versatility. The ability to maneuver in complex environments and the flexibility in configuration empower these systems to efficiently and effectively perform a wide range of tasks across various industries, such as agriculture, construction, delivery, and surveillance \cite{use}.

Recently, deep reinforcement learning (RL) has made impressive progress in robotics applications such as locomotion and manipulation. It has also been successfully applied to drone control and decision-making \cite{hwangbo2017control, belkhale2021model, song2021autonomous, zhang2023learning}, improving the computational efficiency, agility, and robustness of drone controllers. Compared to classic optimization-based methods, RL-based solutions circumvent the need for explicit dynamics modeling and planning and allow us to approach these challenging problems without accurately knowing the underlying dynamics. Moreover, for multi-drone systems, we can further leverage Multi-Agent RL (MARL), which is shown to be effective in addressing the complex coordination problems that arise in multi-agent tasks \cite{vinyals2019grandmaster, berner2019dota, yu2022surprising}. 

Efficient and flexible simulated environments play a central role in RL research. They should allow researchers to conveniently build up the problem of interest and effectively evaluate their algorithms. Extensive efforts have been made to develop simulators and benchmarks for commonly studied robot models like quadrupedals and dexterous arms \cite{rudin2022learning, chen2022towards, hwangbo2018raisim, james2020rlbench}. However, although a range of drone simulators already exists, they suffer from issues such as relatively low sampling efficiency and difficult customization. 

To help better explore the potential of RL in building powerful and intelligent drone systems, we introduce \env, a platform featuring:

\begin{itemize}
    \item Efficiency. Based on Nvidia Isaac Sim \cite{makoviychuk2021isaac, mittal2023orbit}, \env can notably achieve over $10^5$ steps per second in terms of data collection, which is crucial for applying RL-based methods at scale. 
    \item Flexibility. By default, we provide 4 drone models commonly used in related research, along with 4 control modes and 5 sensor modalities, all being easy to extend. We also make it straightforward for users to import their own models and add customized dynamics.
    \item RL-support. \env includes a diverse suite of 10+ single- and multi-agent tasks, presenting different challenges and difficulty levels. The tasks can be easily extended and seamlessly integrated with modern RL libraries.
\end{itemize}

\begin{figure}[t]
    \centering
    \includegraphics[width=\linewidth]{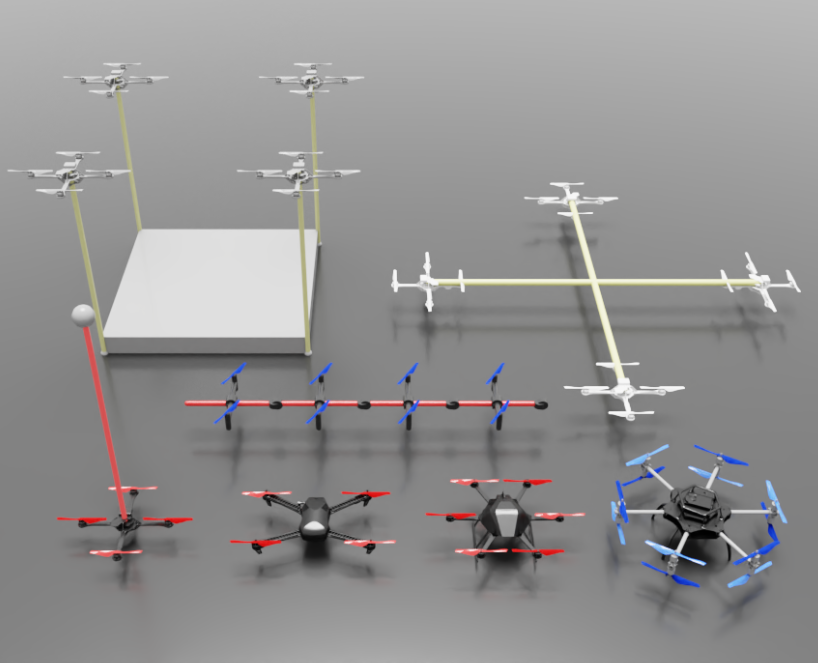}
    \caption{A visualization of the various drone systems in \env, for which we offer highly efficient simulation, reinforcement learning environments, and benchmarking of baselines.}
    \label{fig:enter-label}
\end{figure}

To demonstrate the features and functionalities of \env while also providing some baseline results, we implement and benchmark a spectrum of popular single- and multi-agent RL algorithms on the proposed tasks.  

\begin{figure*}[t]
    \centering
    \includegraphics[width=\linewidth]{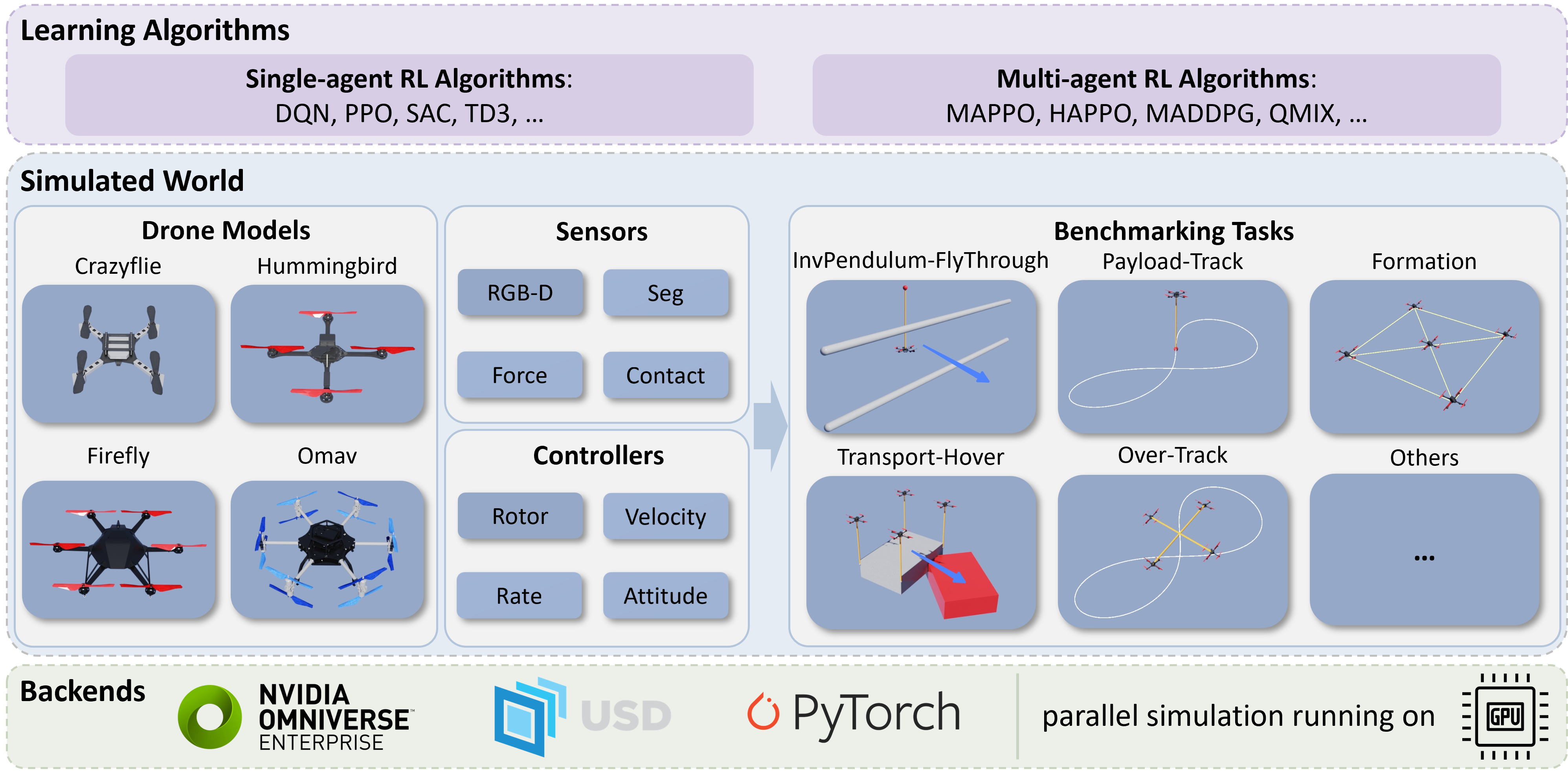}
    \caption{Overview of \env. \env provides a foundational library of various sensors and drone models and offers multiple configurations to form diverse drone systems for multifaceted testing. In addition, \env incorporates several benchmark task logics, enabling the evaluation of the performance of different drone systems across various task objectives. Furthermore, we have implemented and assessed the capabilities of multiple learning algorithms on our benchmark tasks, serving as a baseline for subsequent work.}
    \label{fig:overview}
\end{figure*}

\section{Related Work}

Simulated environments play a crucial role in the RL literature. We highlight the motivation of our work by reviewing the solutions developed out of various considerations, and related research in RL-based control of drones.

\paragraph{Simulated Environments for drones} A common option in the control literature is to use Matlab to perform numerical simulations. This approach enjoys simplicity but has difficulty building complex and realistic tasks and is less friendly to reinforcement learning. Flightmare \cite{song2021flightmare} and Airsim \cite{airsim2017fsr} leverage game engines such as Unity and Unreal Engine that enable visually realistic simulation. Flightmare's efficient C++ implementation can notably achieve $10^6$ FPS but at the cost of being inflexible to extend. Simulators based on the Robot Operating System (ROS) and Gazebo~\cite{koenig2004design} have also been widely used \cite{furrer2016rotors, silano2020crazys} as they provide the ecosystem closest to real-world deployment. For example, RotorS~\cite{furrer2016rotors} provides very fine-grained simulation of sensors and actuators and built-in controllers for the included drone models, enabling sim-to-real transfer of control policies with less effort. However, Gazebo suffers from poor scalability and sample efficiency. Additionally, the working mechanism of ROS makes environment interaction asynchronous, which violates the common implementation practice in RL. To provide an RL-friendly environment, PyBullet-Drones~\cite{panerati2021pybullet-drones} introduced an OpenAI Gym-like environment for quadrotors based on PyBullet physics engine \cite{coumans2016pybullet}. However, it relies on CPU multiprocessing for parallel simulation, which limits its scalability and leads to fewer steps per second.

Our platform aims firstly for efficiency and a friendly workflow for RL. While the highly-parallelized GPU-based simulation ensures a high sampling performance, it is also convenient to customize and extend the environment at Python level and seamlessly work with modern RL libraries such as TorchRL.

\paragraph{Reinforcement learning of drone control} Reinforcement learning is seen as a potential approach for control and decision-making for multi-rotor drones. Prior works explored end-to-end training of visual-motor control policies \cite{zhang2016learning, levine2016end, hwangbo2017control, koch2019reinforcement} to avoid the need for explicit dynamics modeling and hand-engineered control stack. Model-based reinforcement learning can combine learned forward dynamics models with planning methods, such as model predictive control (MPC), and has been investigated in \cite{lambert2019low, belkhale2021model}. Applications to agile drone racing \cite{kaufmann2018deep, song2021autonomous} also demonstrated RL-based policies' ability to cope with highly dynamic tasks, generating smooth and near-time-optimal trajectories in real-time. \cite{kaufmann2022benchmark} benchmarked different choices of action spaces and control levels regarding learning performance and robustness. More recently, \cite{zhang2023learning} trains a single adaptive policy that can control vastly different quadcopters, showing the potential of reinforcement learning in terms of generalization and adaptation capabilities.

To fully uncover what possibilities RL brings to drones, a flexible and versatile platform that supports various research purposes is highly desirable. In light of that, \env aims to be suitable for a range of challenging topics, such as multi-agent coordination, adaptive control, design of modular drones, etc.

\begin{table*}[]
\caption{Comparison between \env and other commonly used simulated environments. In \textbf{Drone Model} columns, \textit{Quad., Hexa., Omni.} stand for quadcopter, hexacopter, and omnidirectional, respectively. In \textbf{Sensor} column, \textit{S} stands for segmentation, \textit{F} stands for force sensors, and \textit{C} stands for contact sensors.}

\label{tab:simulators}
\centering
\resizebox{\textwidth}{!}{
\begin{tabular}{@{}c|cc|cc|ccc|cc|c|c|cc@{}}
    \toprule
     & \multirow{2}{*}{\textbf{Physics Engine}} & \multirow{2}{*}{\textbf{Renderer}} & \multicolumn{2}{c|}{\textbf{Vectorization}} & \multicolumn{3}{c|}{\textbf{Drone Model}} & \multicolumn{2}{c|}{\textbf{Runtime Operation}} & \textbf{Sync. / Steppable} & \multirow{2}{*}{\textbf{Sensor}} & \multicolumn{2}{c}{\textbf{User Interface}} \\
     & & & \textbf{CPU} & \textbf{GPU} &\textbf{Quad.} &\textbf{Hexa.} &\textbf{Omni.} & \textbf{Configuration} & \textbf{Randomization} & \textbf{Physics \& Rendering} & & \textbf{Task Spec} & \textbf{RL API} \\
    \midrule\midrule
    RotorS\cite{furrer2016rotors} & Gazebo-based & OpenGL &\tick &\cross &\tick &\tick &\tick &\cross &\cross & \cross & \textit{IMU, RGBD} & - & -   \\
    Airsim\cite{airsim2017fsr} & PhysX & Unreal Engine &\tick &\cross &\tick &\cross &\cross &\cross &\cross & \cross &\textit{IMU, RGBD, S} & C++\&Python & Single\&Multi. \\
    Flightmare\cite{song2021flightmare} & Flexible & Unity &\tick &\cross &\tick &\cross &\cross &\cross &\tick$^*$ & \tick &\textit{IMU, RGBD, S} & C++ & Single   \\
    PyBullet-Drones\cite{panerati2021pybullet-drones} & Bullet & OpenGL &\tick &\cross &\tick &\cross &\cross &\cross &\cross & \tick &\textit{IMU, RGBD, S} & Python & Single\&Multi.   \\
    FlightGoggles\cite{FlightGoggles} & Flexible & Unity & \tick & \cross & \tick & \cross  & \cross & \cross & \cross & \cross & \textit{IMU, RGBD, S} & C++ & -  \\
    CrazyS\cite{crazys} & Gazebo-based & OpenGL &\tick &\cross &\tick &\tick &\tick &\cross &\cross & \cross & \textit{IMU, RGBD} & - & -   \\
    \midrule
    \textbf{OmniDrones (ours)} & PhysX & Omniverse RTX & \tick & \tick & \tick & \tick & \tick & \tick & \tick &\tick &\textit{IMU, RGBD, S, F, C} & Python & Single+Multi. \\
    \bottomrule
\end{tabular}}

\end{table*}

\section{OmniDrones Platform}


At a high level, \env consists of the following main components: (1) A simulation framework featuring GPU parallelism and flexible extension; (2) Utilities to manipulate and extend the drone models and simulation for various purposes; (3) A suite of benchmark task scenarios built from (1) and (2), serving as examples and starting points for customization.

An overview of \env is presented in \cref{fig:overview}. For comparison, \cref{tab:simulators} contrasts \env with existing drone simulators, highlighting the advantages of our platform. In the following subsections, we describe the details of these components and provide examples to demonstrate the overall workflow.

\subsection{Simulation Framework}

Drones have garnered significant attention from both industry and academia due to their remarkable agility and versatility. For example, a single drone can execute acrobatics or deliver lightweight items independently, while multiple drones can work together to aid in rescue operations in dense forests or transport bulky cargo collaboratively. 

Our simulation framework employs a bottom-up modular design approach to cater to the diverse needs of drone applications. This approach begins by setting up all basic modules of a drone system. Afterward, these modules can be integrated procedurally to simulate complex task scenarios. Following this strategy, our simulation includes a range of basic modules: (1) drone models, (2) sensor stacks, (3) control modes, (4) system configurations, and (5) task specifications. 

Regarding the multi-rotor dynamics, we use the general model given by:
\begin{align}
    & \dot{\mathbf{x}}_W = \mathbf{v}_W & \dot{\mathbf{v}}_W = \mathbf{R}_{WB} \mathbf{f} + \mathbf{g} + \mathbf{F}\\
    & \dot{\mathbf{q}} = \frac{1}{2}\mathbf{q} \otimes \omega & \dot{\omega} = \mathbf{J}^{-1}(\mathbf{\eta} - \omega\times J\omega)
\end{align}
where $\mathbf{x}_W$ and $\mathbf{v}_W$ indicate the position and velocity of the drone in the world frame. $R_{WB}$ is the rotation matrix from the body frame to the world frame. $\mathbf{J}$ is the diagonal inertia matrix, and $\mathbf{g}$ denotes Earth’s gravity. $\mathbf{q}$ is the orientation represented with quaternion, and $\omega$ is the angular velocity. $\otimes$ denotes the quaternion multiplication. $\mathbf{F}$ includes other external forces, e.g., those introduced by the drag and downwash effects. The collective thrust $\mathbf{f}$ and body torque $\eta$ are derived from single rotor thrusts $\mathbf{f}_i$ as:
\begin{align}
    & \mathbf{f} = \sum_i \mathbf{R}_B^{(i)} \mathbf{f}_i \\
    & \eta = \sum_i \mathbf{T}_B^{(i)}\times \mathbf{f}_i + k_i \mathbf{f}_i 
\end{align}
where $\mathbf{T}_B^{(i)}$ and $\mathbf{R}_B^{(i)}$ are the local translation and orientation (tilt) of the $i$-th rotor, $k_i$ the force-to-moment ratio, represented in the body frame.

We offer a range of popular drone models for various applications. We detail four representative drones in this paper, including the \drone{Crazyflie}, a small X-configuration quadrotor; the \drone{Hummingbird}, an H-configuration quadrotor; the \drone{Firefly}, a hexacopter; and the \drone{Omav}, an omnidirectional drone with tiltable rotors. These models vary in size and design, from compact quadrotors to larger omnidirectional drones, each with unique dynamical features. Moreover, our simulator provides an array of sensors such as IMUs, RGB-D cameras, segmentation sensors, force sensors, and contact sensors. This range ensures drones can be easily tailored with the preferred sensor combinations, addressing specific requirements for state estimation and perception. We also implement for most drone models three PD controllers acting on different levels of commands, including position/velocity, body rate, and attitude.

Before delving into the rest of the paper, we outline the primary features of the simulation framework based on the designs mentioned earlier:
\paragraph{Multi-rotor drone dynamics} \env supports drone simulations with variable rotor numbers through a general implementation of drone dynamics, as described above. We also account for external forces in the dynamics, expanding the range of potential tasks.
\paragraph{Parallelism and scalability} Similar to other GPU-based simulators, \env also benefits from the high parallelism and subsequent near-linear scalability of Isaac Sim~\cite{isaacsim}. This enables us to achieve a high-performing policy within a short amount of time.
\paragraph{Physical configuration and rigid dynamics} The physical configuration of a drone model is specified by a Universal Scene Description (USD) file, which can be converted from the URDF format commonly used in Gazebo-based simulations. That means that \env is compatible with drone models that have been used in the community. Notably, with Isaac Sim, it is possible to programmatically modify the physical configuration, e.g., changing its physical properties and assembling with other drones to form multi-drone systems as shown in \cref{fig:overview}.

\subsection{Extending the Drone Models}
\label{sec:config_lib}

Certain applications may require additional payloads to be attached. Also, it might be desirable to create multi-drone systems to cope with tasks beyond a single drone's capability. With the flexible simulation framework, one feature of our platform is the ability to procedurally build and extend a drone system's physical/logical \textbf{configuration} for diverse interests. Notably, they can be generated programmatically from existing drone models and a set of primitives in a highly parameterizable fashion.



Here, we introduce examples of interesting configurations provided in \env. The formed configurations may cause considerable changes in the drone's dynamics and thus present challenges for conventional controller design.

\begin{itemize}
    \item \texttt{Payload \& InvPendulum}: A single drone is connected to a weight through a rigid link. The attached weight will alter and destabilize the drone's dynamics. The arrangement with the payload at the bottom is called \texttt{Payload}, while the arrangement with the payload on top is called \texttt{InvPendulum}.
    \item \texttt{Over-actuated Platform (Over)}: An over-actuated platform consists of multiple drones connected through rigid connections and 2-DoF passive gimbal joints, similar to \cite{su2022downwash}. Each drone functions as a tiltable thrust generator. By coordinating the movements of the drones, it becomes possible to control their positions and attitudes independently, allowing for more complex platform maneuvers.
    \item \texttt{Transport}: A transportation system comprises multiple drones connected by rigid links. This setup allows them to transport loads that exceed the capacity of a single drone. Drones need to engage in coordinated control and collaboration for stable and efficient transportation.
    \item \texttt{Dragon}: A multi-link transformable drone as described in \cite{dragon}. Each link has a dual-rotor gimbal module. The links are connected via 2-Dof joint units sequentially. The ability to transform enables highly agile maneuvers and poses a challenging control problem.
\end{itemize}


\subsection{Randomization}

Since there are unavoidable gaps between the simulated dynamics and reality, randomization is an important and necessary technique for obtaining robust control policies that can be easily transferred and deployed to real-world robots. One particular advantage of having a large number of parallel environments is that we can collect a large volume of diverse data from the randomized distribution, making \env appealing for research regarding Sim2Real transfer and adaptation. We list example factors that users can manipulate in \cref{tab: randomization}.

\begin{table}[h]
\caption{Randomizable Simulation Aspects}
\begin{tabular}{@{}llll@{}}
\toprule
Aspect           & Examples                       & Startup & Runtime \\ \midrule
Physical config. & rigid connection, object scale & \tick        &  \cross      \\
Inertial prop.   & mass, inertia                  & \tick        &  \tick       \\
Rotor param.     & force costant, motor gain      & \tick        &  \tick       \\
External forces  & wind, drag                     & \tick        &  \tick       \\ \bottomrule
\end{tabular}
\label{tab: randomization}
\end{table}

\subsection{Benchmarking Tasks}
\label{sec:bench_tasks}

\begin{figure*}[ht]
    \centering
    \begin{subfigure}[b]{\linewidth}
        \includegraphics[width=\linewidth]{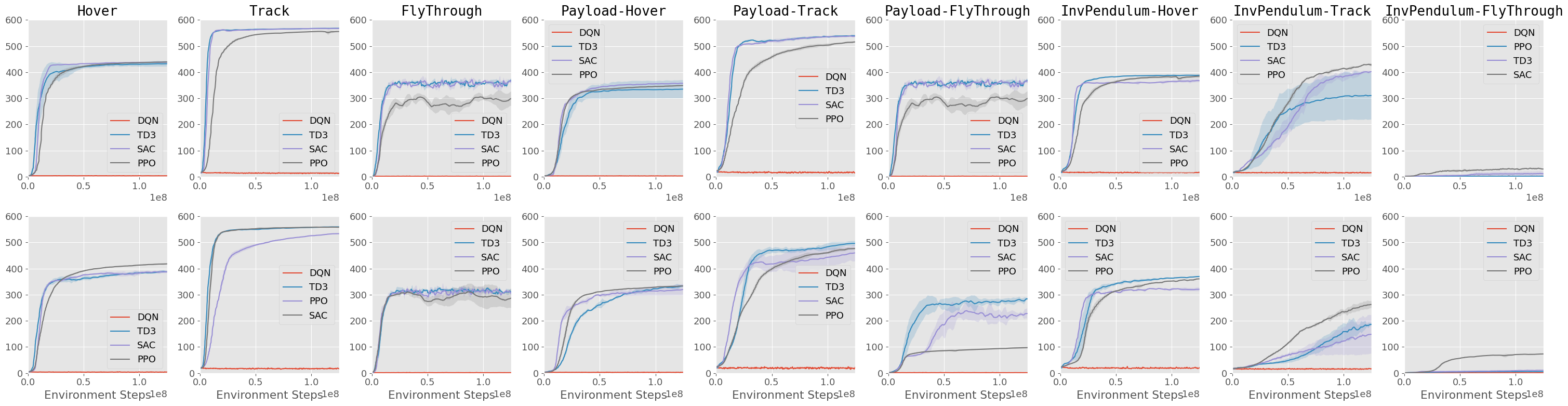}
        \caption{Learning curves of single-agent tasks, with \drone{Hummingbird} (top) and \drone{Firefly} (bottom), respectively.}
        \label{fig:single-agent}
    \end{subfigure}

    \begin{subfigure}[b]{\linewidth}
        \includegraphics[width=\linewidth]{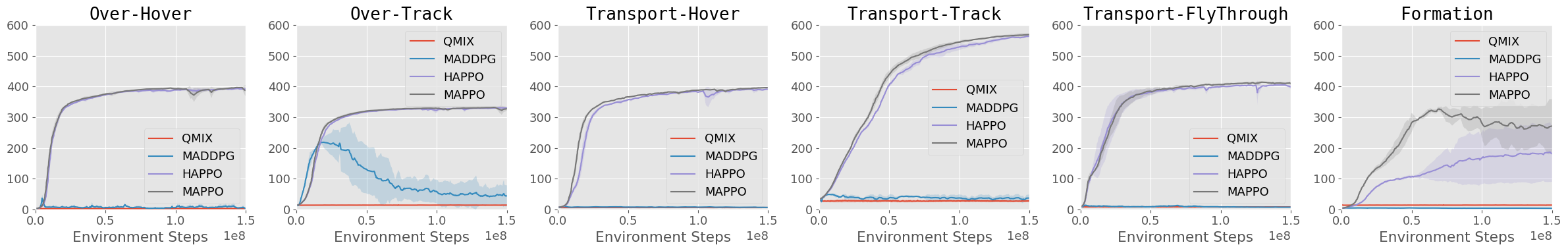}
        \caption{Learning curves of multi-agent tasks.}
        \label{fig:multi-agent}
    \end{subfigure}

    \label{fig:benchmarking}
    \caption{Benchmarking results.}
\end{figure*}

Based on the simulation framework and utilities introduced above, 15 tasks of varying complexity and characteristics are developed for benchmarking. They are formulated as decentralized partially observable Markov Decision Process (Dec-POMDP) \cite{oliehoek2016dec-pomdp}, where partial observability comes from the fact that only a limited part of the system state is known or measured by the sensors and that agents do not have full access to states about each other in a decentralized multi-agent setting. A \textbf{task} specifies the POMDP on top of a certain \textbf{configuration}, similar to DMControl \cite{tassa2018deepmind-control}. For example, \texttt{InvPendulum-Hover} is a task in which the agent (drone) is required to hover an inverted pendulum system introduced before at a desired state. For those that do not have a special configuration, we omit the first part. 

According to their formulations and challenges, we divided the task specifications into categories that each might apply to a set of configurations. Here, we list and introduce several representative examples:

\begin{itemize}
    \item \texttt{Hover}: The drone(s) need to drive the system to reach and maintain a target state. This basic task is simple for most configurations except the inherently unstable ones, e.g., \texttt{InvPendulum}.
    \item \texttt{Track}: The drone(s) are required to track a reference trajectory of states. The ability to (maybe not explicitly) predict how the trajectory would evolve and plan for a longer horizon is needed for accurate tracking.
    \item \texttt{FlyThrough}: The drone(s) must fly the system through certain obstacles in a skillful manner, avoiding any critical collision. The obstacles are placed such that a long sequence of coherent actions is needed. Such a task often challenges the RL algorithm in exploration. 
    \item \texttt{Formation}: A group of drones needs to fly in a specific spatial pattern. This task examines the ability to deal with coordination and credit assignment issues.
\end{itemize}


For detailed specifications on these tasks, please refer to the code.

Generally, each drone observes kinematic information such as relative position, orientation (expressed in quaternions), and linear and angular velocities. Additional sensors can be attached or mounted to RGB-D images if needed. Regarding the action space, the drones are commanded target throttles for each motor, which the underlying motors strive to attain during the control process. 

Additionally, by integrating given with controllers, we can transform the action space to allow for the usage of higher-level control commands. We provide 4 control modes (rotor, velocity, rate, and attitude) for ordinary multi-rotor drones. 


\subsection{Reinforcement Learning with OmniDrones}

It is common in robotics to have RL tasks with complex input and output structures. For example, we might have sensory data from different modalities or want to adopt the teacher-student training scheme where some privileged observation is only visible to a part of the policy. The presence of multiple and potentially heterogeneous agents could introduce further complications. Therefore, to have a flexible interface that conveniently handles tensors in batches, we follow TorchRL's environment specification and use TensorDict as the data carrier, both initially proposed by \cite{bou2023torchrl}. We also provide utilities to transform the observation and action space for common purposes, such as discretizing action space, wrapping a controller, and recording state-action history.

With that, we implement and evaluate various algorithms to provide preliminary results and serve as baselines for subsequent research. They include PPO \cite{schulman2017proximal}, SAC \cite{haarnoja2018soft}, DDPG \cite{fujimoto2018addressing}, and DQN for single-agent tasks and MAPPO \cite{yu2022surprising}, HAPPO \cite{kuba2022trust}, MADDPG \cite{lowe2017multi}, and QMIX \cite{rashid2020monotonic} for multi-agent ones.




\section{Experiments}

Leveraging the simulation framework and benchmark tasks, our platform provides a fair comparative basis for different RL algorithms, serving as a starting point for subsequent investigations. In this section, we showcase the features and functionalities of \env through experiments and evaluate a range of popular RL algorithms on the proposed tasks. In all the following experiments, we use a simulation time step $dt=0.016$, i.e., the control policy operates at around 60Hz.

\subsection{Simulation Performance}

We select a single-agent (\texttt{Track}) and a multi-agent (\texttt{Over-Hover}) task, respectively, to demonstrate the efficient simulation capabilities of our simulator under different numbers of environments.

As shown in \cref{tab:fps}, the efficient PyTorch dynamics implementation and Isaac Sim's parallel simulation capability allow \env to achieve near-linear scalability with over $10^5$ frames per second (FPS) during rollout collection. The results were obtained on a desktop workstation with NVIDIA RTX4090, Isaac Sim 2022.2.0. The control policy is a 3-layer MLP with 256 hidden units per layer implemented with PyTorch. Note that there are additional computations for the observations/rewards and logging logic besides simulation.

\begin{table}[h]
\centering
\caption{Simulation performance (FPS) of \env.}
\begin{tabular}{ccc}
\hline
   \#Envs       &\begin{tabular}[c]{@{}c@{}}\texttt{Track}\\ (1 agents)\end{tabular}        & \begin{tabular}[c]{@{}c@{}}\texttt{Over-Hover}\\ (4 agents)\end{tabular} \\ \hline
1024 Envs & $196074\pm3754$  & $115244\pm1973$                                                     \\
2048 Envs & $385027\pm6688$  & $204556\pm7511$                                                     \\
4096 Envs & $732109\pm10362$ & $310027\pm12233$                                                    \\ \hline
\end{tabular}
\label{tab:fps}
\end{table}

\subsection{Benchmarking RL baselines}
The algorithms are adapted following open-source implementations and modified to be compatible with large-scale training. All runs follow a default set of hyper-parameters without dedicated tuning. Note that the experiments in this part all use direct rotor control.

For single-agent tasks, we evaluate PPO, SAC, DDPG, and DQN using two drone models, namely Hummingbird and Firefly. The two drone models have 4 and 6 action dimensions, respectively, and differ in many inertial properties. For DQN, we discretize the action space by quantizing each dimension into its lower and upper bounds. We train each algorithm in 4096 parallel environments for 125 Million steps. The results are shown in \cref{fig:single-agent}. It can be observed that PPO, SAC, and DDPG are all good baselines for most tasks. However, various failures are observed in some tasks that require substantial exploration to discover the optimal behavior, i.e., \texttt{FlyThrough}. DQN fails to make progress in all tasks.

Notably, PPO-based agents can be trained within 10-20 minutes. On the other hand, SAC and DDPG generally exhibit better sample efficiency. However, they require a longer wall time, since they need a significantly higher number of gradient steps with more data for each update.

For the more challenging multi-agent coordination tasks, we evaluate MAPPO, HAPPO, MADDPG, and QMIX using \drone{Hummingbird}. We train all algorithms for 150M steps. The results are shown in \cref{fig:multi-agent}. The two PPO-based approaches are similar, and both achieve reasonable performance. The failure of MADDPG is potentially due to its exploration strategy being insufficient in multi-agent settings without careful tuning of the exploration noise. To apply the value-decomposition method, QMIX, we discretize the action space as we did for DQN. The results suggest that PPO-based algorithms may serve as strong and robust baselines for obtaining cooperative control policies, which would otherwise require involved analysis of the multi-agent system dynamics.

\subsection{Drone Models and Controllers}

\begin{figure}[h!]
    \centering
    \includegraphics[width=\linewidth]{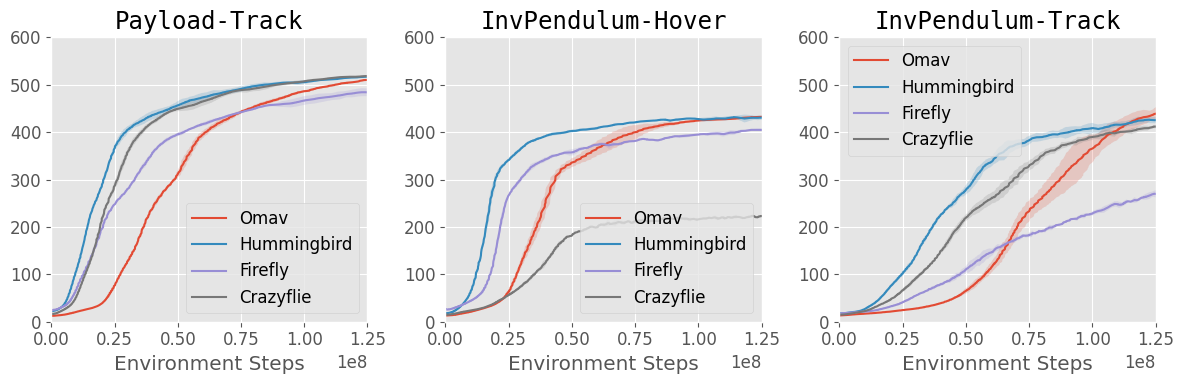}
    \caption{Comparison of different drone models on three selected tasks.}
    \label{fig:drone-model}
\end{figure}

Different drone models render different properties and hence flight performance. It also decides the difficulty of the fundamental aspect of each learning task. The comparison of 4 drone models is shown in \cref{fig:drone-model}. Interestingly, although being the most complex (with 12 rotors and 6 tilt units), \drone{Omav} can be trained to achieve comparable or even better performance on the same budget. This reveals the potential of RL in quickly obtaining a control policy for unusual drone models.

The choice of action space can have a vital impact on the performance and robustness of learned policies \cite{kaufmann2022benchmark}. Considering the usage of a controller as a transform of the action space, we verify this point by comparing the following four approaches using \drone{Firefly} and the implemented controllers: (1) Direct control, i.e., the policy directly commands the target throttle for individual rotors; (2) Velocity control, where the policy outputs the target velocity and yaw angle; (3) Rate control, where the policy outputs the target body rates and collective thrust; (4) Attitude control, where the policy outputs the target attitude and collective thrust. The actions are scaled and shifted to a proper range for each approach. 

\begin{figure}[h!]
    \centering
    \includegraphics[width=\linewidth]{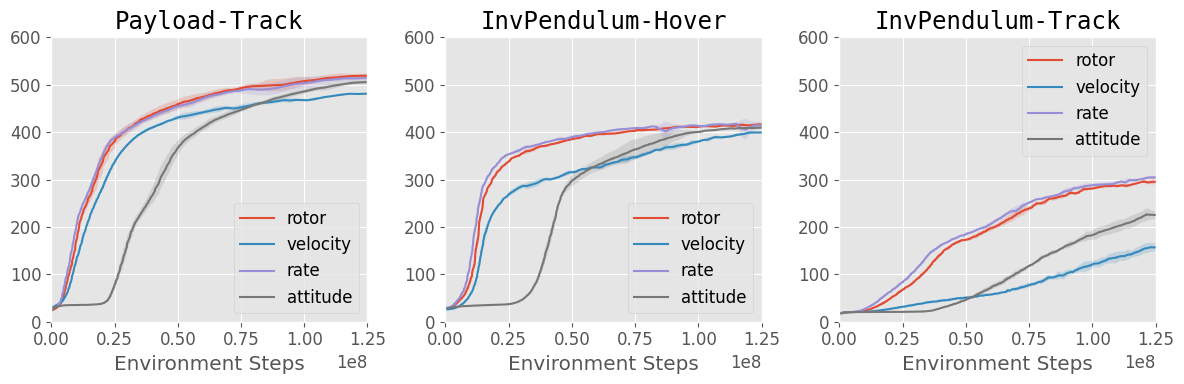}
    \caption{Comparison of different choices of action space.}
    \label{fig:controller}
\end{figure}

As shown in \cref{fig:controller}, direct and rate control consistently gives the best performance, while velocity control appears to be insufficient for tasks that demand more fine-grained control. We remark that tuning the controller parameters tuning and carefully shaping the action space might give a considerable performance boost. Nonetheless, the results suggest that a relatively low-level action space, despite being more subtle to transfer, is still necessary for agile and accurate maneuvers when dynamic changes are present.

\section{Conclusion and Future Work}

In this paper, we presented the \env: a platform for conducting RL research on multirotor drone control. Leveraging the parallel simulation capabilities of more GPUs, \env provides efficient and flexible simulation and a suite of RL tasks for multi-rotor drones. Through experiments, we demonstrate the features of the proposed platform and offer initial results on the tasks. We hope \env serves as a good starting point toward building more powerful drone systems regarding control and system design with reinforcement learning. 

In the future, we will provide long-term support and continue our development to provide utilities for sim-to-real deployment. Current limitations, such as the bottle-necked rendering performance, should be addressed. While this work focuses more on low-level control in an end-to-end setting, more complex and realistic scenarios, and higher-level tasks will be incorporated to complete the picture. 




\section*{ACKNOWLEDGMENT}

This research was supported by National Natural Science Foundation of China (No.62325405, U19B2019, M-0248), Tsinghua University Initiative Scientific Research Program, Tsinghua-Meituan Joint Institute for Digital Life, Beijing National Research Center for Information Science, Technology (BNRist) and Beijing Innovation Center for Future Chips.

The abstractions and implementation of \env was inspired by \textsc{Isaac Orbit} \cite{mittal2023orbit}. Some of the drone models (assets) and controllers are adopted from or heavily based on the RotorS \cite{furrer2016rotors} simulator. We also thank Eric Kuang from NVIDIA for valuable tips on working with the standalone workflow of Isaac Sim.

\bibliographystyle{IEEEtran}
\bibliography{references}

\end{document}